\documentclass[conference]{IEEEtran}

\usepackage{cite}
\usepackage{amsmath,amssymb,amsfonts}
\usepackage{algorithmic}
\usepackage{graphicx}
\usepackage{textcomp}
\usepackage{xcolor}
\usepackage{pgfplots}
\usepackage{pgfplotstable}
\usepgfplotslibrary{groupplots}
\pgfplotsset{compat=1.18}
\usepackage{subcaption}
\usepackage[inkscapelatex=false]{svg}
\usepackage{makecell}
\usepackage{dblfloatfix}

\pgfplotstableread[col sep=comma]{stats/delta_experiment.csv}\datatable

\def\BibTeX{{\rm B\kern-.05em{\sc i\kern-.025em b}\kern-.08em
    T\kern-.1667em\lower.7ex\hbox{E}\kern-.125emX}}
\begin{document}

\title{Silicon Aware Neural Networks\\
}


\author{\IEEEauthorblockN{Sebastian Fieldhouse}
\IEEEauthorblockA{\textit{College of Semiconductor Research} \\
\textit{National Tsing Hua University}\\
}
\and
\IEEEauthorblockN{Kea-Tiong Tang}
\IEEEauthorblockA{\textit{Department of Electrical Engineering } \\
\textit{National Tsing Hua University}\\
}
}

\maketitle

\begin{abstract}
Recent work in the machine learning literature has demonstrated that deep learning can train neural networks made of discrete logic gate functions to perform simple image classification tasks at very high speeds on CPU, GPU and FPGA platforms. By virtue of being formed by discrete logic gates, these Differentiable Logic Gate Networks (DLGNs) lend themselves naturally to implementation in custom silicon - in this work we present a method to map DLGNs in a one-to-one fashion to a digital CMOS standard cell library by converting the trained model to a gate-level netlist. We also propose a novel loss function whereby the DLGN can optimize the area, and indirectly power consumption, of the resulting circuit by minimizing the expected area per neuron based on the area of the standard cells in the target standard cell library. Finally, we also show for the first time an implementation of a DLGN as a silicon circuit in simulation, performing layout of a DLGN in the SkyWater 130nm process as a custom hard macro using a Cadence standard cell library and performing post-layout power analysis. We find that our custom macro can perform classification on MNIST with 97$\%$ accuracy 41.8 million times a second at a power consumption of 83.88 mW.

\end{abstract}


\section{Introduction}

There is a tension that exists when it comes to the design of digital neural network accelerator chips in that deep learning works well with continuous differentiable functions because these enable the convergence of the network's loss topography, but digital circuits at a fundamental level function on discrete mathematics. This means that converting neural networks into a form that runs efficiently on hardware requires many awkward approximations like quantization and activation function simplification. Work by Petersen et al. \cite{petersen2022difflogic} proposes an elegant solution to this problem, whereby instead of using matrix multiplication as the building blocks of the neural network, they are formed by digital logic gate functions instead. This has the benefit of being convertable to very efficient digital logic in CPUs or FPGA to perform inference of the neural network \cite{petersen2024convolutional}. However, as far as the authors are aware, neural networks trained using this technique have not yet been implemented directly in custom silicon. As such, in this paper, we implement a deep differentiable logic gate network (DLGN) directly as standard cells in the open source SkyWater 130nm CMOS process. Additionally, we introduce a learning rule to allow the neural network to optimize the total area of the circuit by feeding information about the area of the standard cells in the target standard cell library to the neural network during training.

\section{Background}

\subsection{Deep Differentiable Logic Gate Networks}
In \cite{petersen2022difflogic} and \cite{petersen2024convolutional}, Petersen et al. showed that neural networks composed purely of logic gates could be trained to perform ultra-efficient image classification on datasets like MNIST and CIFAR-10. Their neural networks learnt to choose one of 16 possible two input logic gate functions, for example NAND, OR, XOR, at each node of the computational graph, however such a method encounters two problems (i) a network of discrete logic gates functions is not differentiable, and (ii) the choice of which logic gate function to use at each node is also a discrete decision and cannot be directly learned via deep learning. As such, Petersen et al. suggest two steps to allow the neural network to become fully differentiable during training and thus learn which logic gate to use at each node of the computational graph to achieve an optimal inference accuracy. Firstly, the discrete logic gate functions must be relaxed to equivalent continuous functions so that they become differentiable, for example logical AND ($a_1 \wedge a_2$) can be relaxed to ($a_1 \cdot a_2$) etc. Secondly, during the training phase the neural network must first learn \textit{which} of the 16 possible logic gates to use. This problem is solved by representing each of the 16 possible logic functions with a probability using the softmax function, which means that the output of this neuron can be represented as an expectation during training. Specifically, during training each node is instantiated with a vector of 16 trainable parameters $\mathbf{z} \in \mathbb{R}^{16}$ representing a weighting of each of the 16 possible 2-input logic gate operations $g_0, \dots, g_{15}$. This means that the output activation of two activations can be represented using an expectation using the equation

\begin{equation}
\begin{aligned}
    f_{\mathbf{z}}(a_1, a_2)
    &= \mathbb{E}_{i \sim \mathcal{S}(\mathbf{z}),\;
      A_1 \sim \mathcal{B}(a_1),\;
      A_2 \sim \mathcal{B}(a_2)}
    \left[ g_i(A_1, A_2) \right] \\
    &= \sum_{i=0}^{15}
      \frac{\exp(z_i)}{\sum_{j} \exp(z_j)}
      \cdot g_i(a_1, a_2).
\end{aligned}
\end{equation}

where $\mathcal{S(\mathbf{z})}$ represents the probability distribution across the 16 options. After training is complete, the neural network is discretized by assigning to each node the logic gate with the highest probability distribution at said node. The DLGN is then formed of $L$ sequential \emph{LogicLayers}, each containing $K$ of the above described neurons, followed by a \emph{GroupSum} aggregation layer at the end of the model that produces class logits. Each neuron $n$ in a LogicLayer receives exactly two inputs $a_n, b_n \in [0, 1]$ selected via a fixed random wiring pattern, and computes a single 2-input logic function, as described above.

\begin{figure}[htbp]
\centerline{\includegraphics[width=\columnwidth]{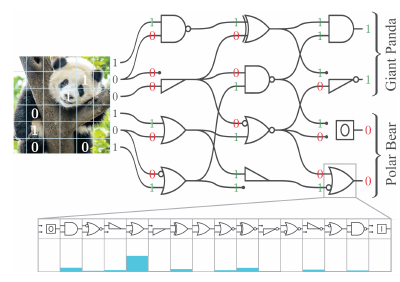}}
\caption{Concept of DLGN for image classification, from \cite{petersen2024convolutional}}
\label{fig}
\end{figure}

\subsection{Standard Cells}
Cell-based IC design uses standard cells to map digital logic function to pre-designed cells designed using the target manufacturing technology. Due to the idiosyncracies of the CMOS process, not all standard cells have the same area, and by extension do not have the same power consumption. For example, implementations of NAND cells will tend to be smaller in area than XOR gates, with NAND cells being implementable with only 4 transistors compared to XOR gates which might require as many as 8-12 transistors, depending on the implementation. As such, designs which use a higher proportion of these cheaper standard cells will also theoretically have better PPA. Standard cell libraries tend to be custom drawn for each individual process; different companies will also use their own standard cell libraries, and as such not all standard cells will be available in all standard cell libraries. However, one certainty is that the standard cells representing different logic gate functions will have different areas so there will always be a certain amount of optimization that could be performed if the circuit to be implemented was somewhat indifferent to which specific logic gate functions are used, as is the case with deep differentiable logic gate networks. It should be noted that standard cell libraries tend to be the intellectual property of private companies, and as such it is mostly prohibited to publically discuss or publish information relating to them. As such, in this work we used an open source standard cell library provided by Cadence for the SkyWater 130nm process for our experimentation.

\subsection{SkyWater 130nm pdk}
SkyWater 130nm is a mature 1P5M process offered by the company SkyWater Technology - the SkyWater 130~nm process design kit (PDK) is an experimental open-source PDK for CMOS integrated circuit design, enabling transparent and reproducible hardware research workflows \cite{edwards2020skywater} \cite{skywater2020pdk}. Developed through the SkyWater and Google open-source program, it provides technology files, design rules, device models, and reference standard-cell libraries required for full custom and digital ASIC development. In this work we use a standard cell library provided by Cadence under an open-source license for the SkyWater 130nm process. The reason for this is that the Cadence standard cell library is written in a format that makes it integrate well with Cadence tools, which were used throughout this work. 

\section{Methodology}
Firstly, since we are targeting the SkyWater 130nm process to implement our deep differentiable logic gate network in, we map all of the 16 possible 2-input logic gate functions to standard cells in the Cadence-provided library. For consistency, we only use standard cells of low driving strength. However, some of the logic functions do not have a one-to-one mapping to the standard cells in our library, so we map them to a combination of standard cells, as described in Table~\ref{tab:logic_gate_mapping} and then estimate their area as being the sum of the two standard cells used to create this logic function.

\begin{table}[htbp]
\caption{Mapping of logic gate functions to SkyWater 130\,nm Cadence standard cells}
\label{tab:logic_gate_mapping}
\begin{center}
\renewcommand{\arraystretch}{1.3}
\begin{tabular}{|c|c|c|c|}
\hline
\textbf{Function} & \textbf{Truth Table} & \textbf{Available Cell(s)} & \textbf{Area ($\mu$m\textsuperscript{2})} \\
\hline
$0$           & 0000 & TIELO          &  5.713 \\
\hline
$A \cdot B$             & 0001 & AND2X1         &  9.522 \\
\hline
$A \cdot \overline{B}$  & 0010 & INVX1 + NOR2X1   & 13.331 \\
\hline
$A$                 & 0011 & BUFX2      &  7.618 \\
\hline
$\overline{A} \cdot B$   & 0100 & INVX1 + NOR2X1 & 13.331 \\
\hline
$B$       & 0101 & BUFX2      &  7.618 \\
\hline
$A \oplus B$           & 0110 & XOR2X1         & 15.235 \\
\hline
$A + B$            & 0111 & OR2X1        &  9.522 \\
\hline
$\overline{A + B}$           & 1000 & NOR2X1             &  7.618 \\
\hline
$\overline{A \oplus B}$    & 1001 & XNOR2X1    & 15.235 \\
\hline
$\overline{B}$        & 1010 & INVX1       &  5.713 \\
\hline
$\overline{B} + A$   & 1011 & INVX1 + NAND2X1   & 13.331 \\
\hline
$\overline{A}$     & 1100 & INVX1     &  5.713 \\
\hline
$\overline{A} + B$  & 1101 & INVX1 + NAND2X1   & 13.331 \\
\hline
$\overline{A \cdot B}$   & 1110 & NAND2X1    &  7.618 \\
\hline
$1$        & 1111 & TIEHI     &  5.713 \\
\hline
\end{tabular}
\end{center}
\end{table}

\subsection{Loss Function for Area}
In order to optimize the area utilization of the neural network, we introduce an area-aware loss function to minimize the area cost associated with the current weights learnt by the neural network at any given training step. We do this by calculating an expected area for the total DLGN - since the softmax gate probabilities $\mathbf{p}_n$ represent the likelihood of each gate type being selected, the \emph{expected area} of neuron $n$ is:

\begin{equation}
    \mathbb{E}[\text{area}_n] = \mathbf{p}_n^\top \mathbf{A} = \sum_{i=0}^{15} p_{n,i} \cdot A_i
\end{equation}

This expectation is a differentiable function of the learnable weights $\mathbf{z}_n$ through the softmax, enabling gradient-based optimization of hardware area.

The area loss is computed as the mean expected area across all $N$ neurons in the network:
\begin{equation}
    \mathcal{L}_{\text{area}} = \frac{1}{N} \sum_{n=1}^{N} \mathbb{E}[\text{area}_n]
    \label{eq:area_loss}
\end{equation}
Normalizing by the total neuron count $N$ makes $\mathcal{L}_{\text{area}}$ interpretable as the average expected area per neuron (in $\mu\text{m}^2$) and ensures that the penalty coefficient remains meaningful across architectures of varying width and depth.

We combine our area loss term with the classification cross-entropy loss:
\begin{equation}
    \mathcal{L}_{\text{total}} = \mathcal{L}_{\text{CE}} + \delta \cdot \mathcal{L}_{\text{area}}
    \label{eq:total_loss}
\end{equation}
where $\delta \geq 0$ is a hyperparameter controlling the trade-off between classification accuracy and circuit area. We then empirically choose $\delta$ to cause the model to minimize area without negatively impacting classification accuracy.

To validate our proposed concept, we train DLGNs on the MNIST and CIFAR-10 datasets with and without our area-aware compound loss, using the areas from Table~\ref{tab:logic_gate_mapping} to calculate the area loss. Once the model is trained and discretized, we map it directly to a gate level netlist using standard cells in our standard cell library. The \emph{LogicLayers} are mapped to the standard cells described in Table~\ref{tab:logic_gate_mapping} and ADDHX1 and ADDFX1 standard cells are used to create a popcount binary adder tree to perform the summation of the layer outputs for \emph{GroupSum}. We found that the model architecture proposed in previous works of 6 layers of 64,000 neurons each was difficult to route with the limited routing resources of the 1P5M SkyWater 130nm process, so we choose a more narrow architecture with 18 layers of 4,000 neurons each and find that this can maintain accuracy whilst being more tractable to route in SkyWater 130nm. We note that modern CMOS processes have more routing resources so this may not be an issue at more advanced process nodes. Finally, we perform functional verification via gate-level simulation of the netlist with Xcelium, then we layout the design as a fully combination hard macro with Innovus.

\section{Results}
\subsection{Loss Function Study}
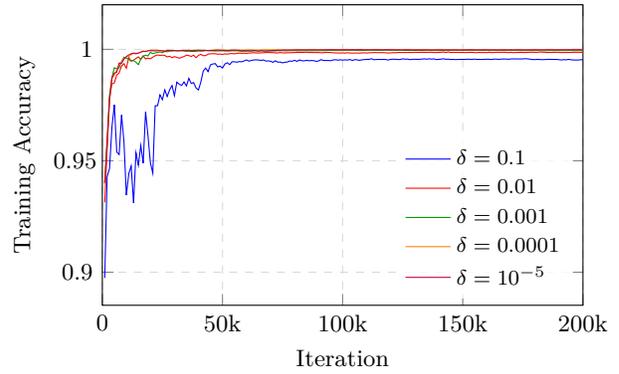
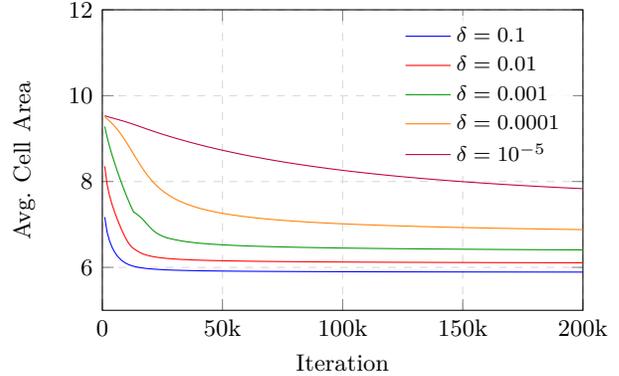
\begin{figure}[t]
  \centering
 
  \begin{subfigure}[t]{0.9\columnwidth}
    \centering
    \hspace{-2em}
    \begin{tikzpicture}
      \begin{axis}[
        width=\linewidth,
        height=0.7\linewidth,
        xlabel={Iteration},
        ylabel={Training Accuracy},
        xmin=0, xmax=200000,
        ymax=1.02,
        ytick={0.85,0.90,0.95,1.00},
        yticklabel style={text width=1.5em, align=right},
        scaled x ticks=false,
        xtick={0,50000,100000,150000,200000},
        xticklabels={0,50k,100k,150k,200k},
        legend style={
          at={(0.98,0.02)},
          anchor=south east,
          font=\footnotesize,
          cells={anchor=west},
          draw=none,
          fill opacity=0.8,
          text opacity=1,
        },
        grid=major,
        grid style={dashed, gray!30},
        line width=0.4pt,
        tick label style={font=\small},
        label style={font=\small},
      ]
 
        \addplot[color=blue, mark=none, filter discard warning=false,
          restrict expr to domain={\thisrow{delta}}{0.099:0.101}]
          table[x=iteration, y=train_acc_eval_mode]{\datatable};
        \addlegendentry{$\delta=0.1$}
 
        \addplot[color=red, mark=none, filter discard warning=false,
          restrict expr to domain={\thisrow{delta}}{0.0099:0.0101}]
          table[x=iteration, y=train_acc_eval_mode]{\datatable};
        \addlegendentry{$\delta=0.01$}
 
        \addplot[color=green!60!black, mark=none, filter discard warning=false,
          restrict expr to domain={\thisrow{delta}}{0.00099:0.00101}]
          table[x=iteration, y=train_acc_eval_mode]{\datatable};
        \addlegendentry{$\delta=0.001$}
 
        \addplot[color=orange, mark=none, filter discard warning=false,
          restrict expr to domain={\thisrow{delta}}{0.000099:0.000101}]
          table[x=iteration, y=train_acc_eval_mode]{\datatable};
        \addlegendentry{$\delta=0.0001$}
 
        \addplot[color=purple, mark=none, filter discard warning=false,
          restrict expr to domain={\thisrow{delta}}{0.0000099:0.0000101}]
          table[x=iteration, y=train_acc_eval_mode]{\datatable};
        \addlegendentry{$\delta=10^{-5}$}
 
      \end{axis}
    \end{tikzpicture}
    \caption{Training accuracy over iterations.}
    \label{fig:train_acc}
  \end{subfigure}

  \begin{subfigure}[t]{0.9\columnwidth}
    \centering
        \hspace{-2em}
    \begin{tikzpicture}
      \begin{axis}[
        width=\linewidth,
        height=0.7\linewidth,
        xlabel={Iteration},
        ylabel={Avg.\ Cell Area},
        xmin=0, xmax=200000,
        ymin=5, ymax=12,
        yticklabel style={text width=1.5em, align=right},
        scaled x ticks=false,
        xtick={0,50000,100000,150000,200000},
        xticklabels={0,50k,100k,150k,200k},
        legend style={
          at={(0.98,0.98)},
          anchor=north east,
          font=\footnotesize,
          cells={anchor=west},
          draw=none,
          fill opacity=0.8,
          text opacity=1,
        },
        grid=major,
        grid style={dashed, gray!30},
        line width=0.4pt,
        tick label style={font=\small},
        label style={font=\small},
      ]
 
        \addplot[color=blue, mark=none, filter discard warning=false,
          restrict expr to domain={\thisrow{delta}}{0.099:0.101}]
          table[x=iteration, y=avg_area_per_neuron]{\datatable};
        \addlegendentry{$\delta=0.1$}
 
        \addplot[color=red, mark=none, filter discard warning=false,
          restrict expr to domain={\thisrow{delta}}{0.0099:0.0101}]
          table[x=iteration, y=avg_area_per_neuron]{\datatable};
        \addlegendentry{$\delta=0.01$}
 
        \addplot[color=green!60!black, mark=none, filter discard warning=false,
          restrict expr to domain={\thisrow{delta}}{0.00099:0.00101}]
          table[x=iteration, y=avg_area_per_neuron]{\datatable};
        \addlegendentry{$\delta=0.001$}
 
        \addplot[color=orange, mark=none, filter discard warning=false,
          restrict expr to domain={\thisrow{delta}}{0.000099:0.000101}]
          table[x=iteration, y=avg_area_per_neuron]{\datatable};
        \addlegendentry{$\delta=0.0001$}
 
        \addplot[color=purple, mark=none, filter discard warning=false,
          restrict expr to domain={\thisrow{delta}}{0.0000099:0.0000101}]
          table[x=iteration, y=avg_area_per_neuron]{\datatable};
        \addlegendentry{$\delta=10^{-5}$}
 
      \end{axis}
    \end{tikzpicture}
    \caption{Average cell area over iterations.}
    \label{fig:avg_area}
  \end{subfigure}
 
  \caption{Training dynamics across five values of $\delta$. Models are initialized with random unique connections, 6 LogicLayers each with 64,000 neurons
           (a)~Training accuracy and (b)~average cell area measured every
           1,000 iterations over 200,000 training steps.}
  \label{fig:delta_experiments}
\end{figure}

First, we run an experiment to find a value of $\delta$ that lets the model minimize the area of the circuit without negatively impacting classification accuracy, which we present in figure\ref{fig:delta_experiments}. We empirically find that a $\delta$ of 0.01 efficiently minimizes the area of the DLGN without having a large impact on accuracy.

\subsection{MNIST and CIFAR-10}

Next, we present implementations of the deep differentiable logic gate network predicted for MNIST and CIFAR-10 with and without the loss function proposed.

\begin{table}[htbp]
\caption{Area Loss Performance}
\centering
\begin{tabular}{|c|c|c|c|}
\hline
\textbf{Model} & \textbf{Accuracy} & \textbf{Avg Area} & \textbf{Total} \\
   &   & \textbf{per Neuron ($\mu\mathrm{m}^2$)} & \textbf{Area ($\mu\mathrm{m}^2$)} \\
\hline
\makecell{MNIST DLGN\\baseline} & 98.04\% & 9.380 & 3,001,592 \\
\makecell{MNIST DLGN\\with area loss} & 97.66\% & 6.107 & 1,954,215 \\
\makecell{CIFAR-10 DLGN\\baseline} & 60.07\% & 9.994 & 12,791,845 \\
\makecell{CIFAR-10 DLGN\\with area loss} & 58.82\% & 7.514 & 9,617,272 \\
\hline
\end{tabular}
\label{tab:area_loss_performance}
\end{table}

\begin{table*}[t]
\caption{Performance Analysis}
\begin{center}
\renewcommand{\arraystretch}{1.3}
\begin{tabular}{|c|c|c|c|c|c|c|c|}
\hline
 & \textbf{\textit{This Work}} 
 & \makecell{\textit{NeurIPS'24}\\\cite{petersen2024convolutional}} 
 & \makecell{\textit{arXiv'25}\\\cite{svein2025tsetlin}} 
 & \makecell{\textit{FPGA'17}\\\cite{umuroglu2017finn}} 
 & \makecell{\textit{Datasheet}\\\cite{max78000}} 
 & \makecell{\textit{CICC'21}\\\cite{lu2021ternary}} 
 & \makecell{\textit{JSSC'19}\\\cite{bankman2019always}} \\
\hline
Accuracy & 97.49\% & 99.23\% & 97.42\% & 95.8\% & 99.6\% & 97.1\% & 97.9\% \\
\hline
Latency & 15\,ns & $<$10\,ns & 16.6\,$\mu$s & 310\,ns & 360\,$\mu$s & -- & 4.2\,ms \\
\hline
\makecell{Equivalent 16\,nm\\Latency} & 4.2\,ns & -- & -- & -- & -- & -- & -- \\
\hline
\makecell{Energy per\\Inference} & 352\,pJ & -- & 8.6\,nJ & -- & 1.1\,$\mu$J & 0.18\,$\mu$J & -- \\
\hline
\makecell{Equivalent 16\,nm\\Energy per Inference} & 69\,pJ & -- & -- & -- & -- & -- & -- \\
\hline
\makecell{Implementation\\Technology} & \makecell{130\,nm\\CMOS} & \makecell{FPGA\\(XC7Z045)} & \makecell{65\,nm\\CMOS} & \makecell{FPGA\\(Zynq ZC706)} & \makecell{40\,nm\\CMOS} & \makecell{28\,nm\\CMOS} & \makecell{28\,nm\\CMOS} \\
\hline
\end{tabular}
\label{tab:performance_comparison}
\end{center}
\end{table*}

\begin{figure}[htbp]
\centerline{\includegraphics[width=0.8\columnwidth]{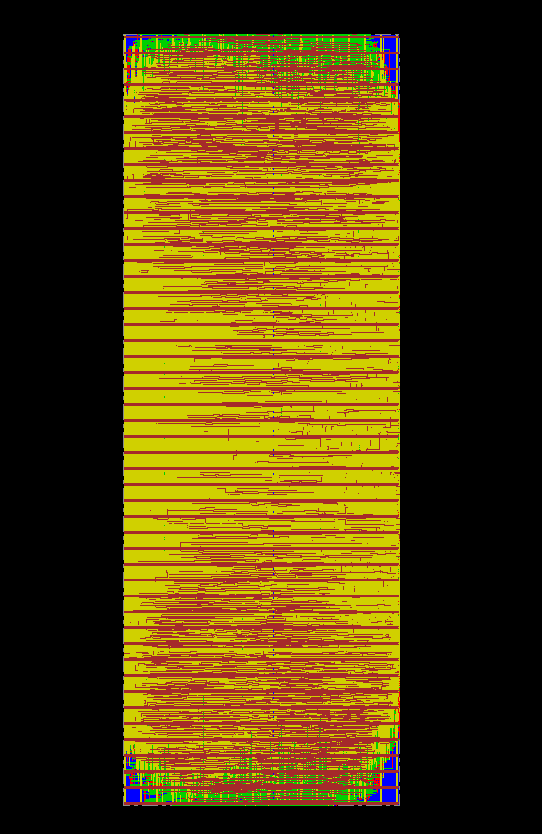}}
\caption{DLGN laid out in SkyWater 130nm as a hard macro}
\label{fig:macro_layout}
\end{figure}

\subsection{Latency and Power Analysis}
Finally, we perform the layout of a hard macro for the MNIST differentiable logic gate network trained with our custom loss function, shown in Fig.~\ref{fig:macro_layout}.

We then use this hard macro to perform post-layout power analysis, from which we estimate energy per inference. We find that the worst-case critical path delay of our circuit is 23.9\,ns, giving a maximum possible throughput of 41.8 million inferences per second. Assuming an input switching activity factor of 0.7, our macro has a power consumption of 83.88\,mW, resulting in an energy per inference of 2.0\,nJ.

We also estimate the equivalent latency and energy of our macro in a 16\,nm process for the purpose of comparison with other works. To do so, we employ the fan-out-of-four (FO4) inverter delay as a technology-independent measure of logical depth~\cite{Stillmaker2017}. The simulated FO4 delay for 130\,nm bulk CMOS is 34.7\,ps and for 16\,nm high-performance multi-gate (FinFET) is 6.12\,ps at their respective nominal supply voltages~\cite{Stillmaker2017}. Our 23.9\,ns critical path thus corresponds to a logical depth of approximately 689 FO4 inverter delays. Applying the 16\,nm FO4 delay, we estimate a scaled latency of approximately 4.2\,ns. For energy scaling, we apply the per-inverter energy ratio from~\cite{Stillmaker2017}, where the simulated FO4 inverter energy is 5.20\,fJ at 130\,nm and 0.179\,fJ at 16\,nm, yielding a scaling factor of 0.034. Applying this to our measured energy per inference gives an estimated 16\,nm energy of approximately 69\,pJ. 

\section{Conclusion}
In this work we have presented the first mapping of a DLGN to silicon in simulation. Future work could perform tapeout and validate power consumption.

\section*{Acknowledgment}

This work was supported by the National Science and Technology Council, Taiwan, under contract no. NSTC 114-2223-E-007-015 and NSTC 114-2622-8-007-005.

\bibliographystyle{IEEEtran}
\bibliography{references}

@inproceedings{petersen2022difflogic,
  title={{Deep Differentiable Logic Gate Networks}},
  author={Petersen, Felix and Borgelt, Christian and Kuehne, Hilde and Deussen, Oliver},
  booktitle={Conference on Neural Information Processing Systems (NeurIPS)},
  year={2022}
}

@inproceedings{petersen2024convolutional,
  author    = {Petersen, Felix and Kuehne, Hilde and Borgelt, Christian and Welzel, Julian and Ermon, Stefano},
  title     = {Convolutional Differentiable Logic Gate Networks},
  booktitle = {Advances in Neural Information Processing Systems (NeurIPS)},
  volume    = {37},
  year      = {2024}
}

@inproceedings{edwards2020skywater,
  author    = {Edwards, R. Timothy},
  title     = {{Google/SkyWater} and the Promise of the Open {PDK}},
  booktitle = {Workshop on Open-Source EDA Technology (WOSET)},
  year      = {2020},
  url       = {https://woset-workshop.github.io/PDFs/2020/a03.pdf}
}

@misc{skywater2020pdk,
  author       = {{SkyWater Technology} and {Google}},
  title        = {{SkyWater SKY130 Open Source PDK}},
  year         = {2020},
  howpublished = {\url{https://github.com/google/skywater-pdk}},
  note         = {Accessed: 2026-03-21}
}

@inproceedings{svein2025tsetlin,
  author    = {Svein Anders Tunheim and others},
  title     = {An All-digital 65-nm Tsetlin Machine Image Classification Accelerator with 8.6 nJ per MNIST Frame at 60.3k Frames per Second},
  year      = {2025},
  note      = {arXiv:2501.19347}
}

@inproceedings{umuroglu2017finn,
  author    = {Yaman Umuroglu and Nicholas J. Fraser and Giulio Gambardella and Michaela Blott and Philip Leong and Magnus Jahre and Kees Vissers},
  title     = {{FINN}: A Framework for Fast, Scalable Binarized Neural Network Inference},
  booktitle = {Proceedings of the 2017 ACM/SIGDA International Symposium on Field-Programmable Gate Arrays},
  year      = {2017},
  pages     = {65--74}
}

@misc{max78000,
  author       = {{Analog Devices}},
  title        = {{MAX78000}: Artificial Intelligence Microcontroller with Ultra-Low-Power Convolutional Neural Network Accelerator},
  year         = {2021},
  howpublished = {Datasheet},
  url          = {https://www.analog.com/en/products/max78000.html}
}

@inproceedings{lu2021ternary,
  author    = {Xiyuan Tang and others},
  title     = {A 0.18$\mu$J/97.1\%-Accuracy Mixed-Signal Ternary {CNN} 
               Accelerator with All On-Chip Processing},
  booktitle = {IEEE Custom Integrated Circuits Conference (CICC)},
  year      = {2021}
}

@article{bankman2019always,
  author  = {Daniel Bankman and Lita Yang and Bert Moons and 
             Mario Verhelst and Boris Murmann},
  title   = {An Always-On 3.8$\mu$J/86\% {CIFAR-10} Mixed-Signal Binary 
             {CNN} Processor With All Memory on Chip in 28-nm {CMOS}},
  journal = {IEEE Journal of Solid-State Circuits},
  volume  = {54},
  number  = {1},
  pages   = {158--172},
  year    = {2019}
}

@article{Stillmaker2017,
  author  = {Aaron Stillmaker and Bevan M. Baas},
  title   = {Scaling equations for the accurate prediction of {CMOS} device performance from 180 nm to 7 nm},
  journal = {Integration, the {VLSI} Journal},
  volume  = {58},
  pages   = {74--81},
  year    = {2017},
  doi     = {10.1016/j.vlsi.2017.02.002}
}

\end{document}